\documentclass[pdflatex,sn-mathphys-num]{sn-jnl}

\usepackage{graphicx}%
\usepackage{multirow}%
\usepackage{amsmath,amssymb,amsfonts}%
\usepackage{amsthm}%
\usepackage{mathrsfs}%
\usepackage[title]{appendix}%
\usepackage{xcolor}%
\usepackage{textcomp}%
\usepackage{manyfoot}%
\usepackage{booktabs}%
\usepackage{algorithm}%
\usepackage{algorithmicx}%
\usepackage{algpseudocode}%
\usepackage{listings}%
\usepackage{float}%
\usepackage{hyperref}
\usepackage{caption} 

\theoremstyle{thmstyleone}%
\theoremstyle{thmstyletwo}%

\theoremstyle{thmstylethree}%

\begin{document}

\title[Article Title]{A New Technique for AI Explainability using Feature Association Map}

\author[1]{\fnm{Sayantani} \sur{Ghosh}}\email{sayantani31g@gmail.com}
\author[2]{\fnm{Amit Kumar} \sur{Das}}\email{amitkrdas.kol@gmail.com}
\author[3]{\fnm{Amlan} \sur{Chakrabarti}}\email{acakcs@caluniv.ac.in}

\affil[1]{\orgname{DBS Bank}}
\affil[2]{\orgname{Institute of Engineering \& Management}}
\affil[1,2,3]{\orgname{University of Calcutta}}

\abstract{Lack of transparency in AI systems poses challenges in critical real-life applications. It is important to be able to explain the decisions of an AI system to ensure trust on the system. Explainable AI (XAI) algorithms play a vital role in achieving this objective. In this paper, we are proposing a new algorithm for Explaining AI systems, FAMeX (\underline{F}eature \underline{A}ssociation \underline{M}ap based \underline{eX}plainability). The proposed algorithm is based on a graph-theoretic formulation of the feature set termed as Feature Association Map (FAM). The foundation of the modelling is based on association between features. The proposed FAMeX algorithm has been found to be better than the competing XAI algorithms - Permutation Feature Importance (PFI) and SHapley Additive exPlanations (SHAP). Experiments conducted with eight benchmark algorithms show that FAMeX is able to gauge feature importance in the context of classification better than the competing algorithms. This definitely shows that FAMeX is a promising algorithm in explaining the predictions from an AI system.}




\maketitle

\section{Introduction}\label{sec1}

For the last few decades, artificial intelligence (AI) based systems have been adopted in multiple facets of daily life. Whether we look at frequent applications like weather forecasting, prediction of stock prices, or we look at more advanced applications like early detection of diseases from medical images, analyse sentiment of texts, etc., the role of AI systems is quite predominant. Major applications of AI systems, as we can see, are in the prediction of events, for which certain complex learning algorithms are employed. These complex learning algorithms can give superior, almost close to accurate, predictions. 

The success of AI systems has also brought forward some deep-lying challenges. When we look at sophisticated AI algorithms, that albeit includes the complex learning algorithms, it is sometimes very difficult to understand how the algorithms are making decisions. It is sometimes impossible to understand whether the decision made by the AI system is based on scientific facts or spurious coincidences. This is fine when we work with non-critical systems like one which recommends books, movies or music to users. But when we go for critical implementations of AI like detection of diseases, forecasting weather, or predicting stock prices, the consequences might be fatal. So, for critical AI based decision systems, the opacity of the black-box AI system needs to be broken and an explanation of the decisions taken by the system should be evident. To foster trust within people, it is essential to conserve reliability, accountability, etc. to maintain transparency between computational systems and human beings [1].

To make an AI system transparent and to explain why a particular algorithm takes a specific decision, the concept of explainability of AI systems has evolved. In the last few years, a lot of research has been done by the AI research community which is focused on establishing transparency of AI systems. Explainable AI (or popularly termed as XAI) algorithms [2, 3, 4] are intended to come up with certain explanations or understanding about how a particular AI system is taking decisions. XAI algorithms have seen a new apex in recent years [5, 6] because conventional forecasting of AI decision-making systems need to be complimented with human-understandable rationale. This has been further necessitated by the fact that the more predictive a system is, the less is its interpretability [7].

Broadly, XAI algorithms are based on few key philosophies. The first one is based on perceptible interpretation. These algorithms try to explain the decisions by AI systems using human perception-based intuitions. The other approach is to explain the AI systems by doing mathematical modelling of the system. A popular approach for explainability is based on feature importance. In this approach the XAI algorithm tries to represent feature importance in the form of ranking of features. There are a few algorithms proposed by the researchers to come up with the feature importance in context of decision-making of an AI system [12, 13]. But they have not done a holistic analysis of the critical aspects of the features - relevance and redundancy. A graph-based formulation of the feature set, termed as Feature Association Map (FAM), proposed in our earlier works [27, 28] has been able to successfully model the aspects of features namely feature relevance and redundancy. We envisage that the same formulation of FAM can also be used to interpret feature importance in context of understanding a prediction done by an AI system in a more holistic way.

The novelty of the algorithm that we have proposed in this paper is that it looks at the feature importance in context of both relevance and redundancy. Hence, it is able to evaluate the criticality of contribution of the feature in the decision-making process of the AI system. FAM, which is a visual modelling of the feature set, has been used while coming up with the feature importance.

The remaining part of the paper has been organized in the following sections: 
\begin{itemize}
\item Section II presents some of the existing methodologies of explainable AI which are relevant to our contribution.
\item Section III highlights the underlying conceptual framework used in the proposed work.
\item Section IV presents the proposed FAMeX algorithm in details.
\item Extensive experimental results have been reported in Section V. This section also contains the comparative study and validation of FAMeX.
\item The work has been summarized in the form of conclusion in Section VI.
\end{itemize}

\section{Related Works}\label{sec2}

XAI techniques are adopted by users to establish trust on AI solutions. LIME [8] uses local models to explain its prediction, Deep LIFT [9] explains the output with the help of backpropagation through the neurons of a network, Partial Dependence Plot [10] explains how the model’s prediction partially depends on values of the input variables of interest.

Permutation Feature Importance (PFI) [11] calculates the increase or decrease of the model’s prediction error after permuting i.e. removing a feature. This gives a measure of the feature importance. Among all the model agnostic local explanation techniques the most popular unified approach is SHAP typified for SHapley Additive exPlanations [13, 14, 15, 16]. It estimates the importance of each feature based on its marginal contribution. SHAP uses the concept of game theory that calculates the respective contribution of the players in a group (coalition). To interpret the learning models it estimates the contribution of each feature to a model’s decision. In a work [17], SHAP is used to interpret and analyze the features obtained by the occurrence of accidents in Chicago Metropolitan Expressways.

\section{Conceptual Framework}\label{sec3}

\subsubsection{Feature Similarity}
In a set of features, the degree of similarity between the features represents one aspect of feature association [2][25][26]. Similarity expresses the redundancy between the set of features drawing a clear inference from the measured values. To measure the similarity between the features, Pearson’s correlation coefficient is extensively used. For a pair of random variables X and Y, correlation coefficient r is defined by the equation 1.
\begin{equation}
r = \frac{cov(X, Y)}{\sqrt{var(X)var(Y)}}
\end{equation}
where
\begin{equation}
cov(X, Y) = \frac{1}{n}\sum{({X}_i - \overline{X})({Y}_i - \overline{Y})}
\end{equation}
and
\begin{eqnarray}
var(X) = \frac{1}{n}\sum{{({X}_i - \overline{X})}^2} \\
var(Y) = \frac{1}{n}\sum{{({Y}_i - \overline{Y})}^2}
\end{eqnarray}

Correlation coefficients range between -1 and +1, where a maximum of linear correlation is indicated by + 1, minimum correlation by -1 and no correlation is indicated by 0. Range of values between 0.68 and 1.00 is considered as high. Moreover, values larger than 0.9 indicate very high correlation [21].\\
\subsubsection{ Feature Relevance}
Feature relevance is a measure of how much information a feature contributes in deciding the value of the target variable. Mutual Information (MI) using entropy is used as a measure of feature relevance. MI between two random variables is a non-negative value, which measures the dependency between the variables. It is equal to zero if and only if two random variables are independent, and higher values mean higher dependency [9]. Mutual information between a feature \textit{f} and the target variable \textit{C} can be calculated as shown in equation 5.
\begin{equation}
MI(C,~f)~=~H(C)~+~H(f)~-H(C,~f)
\end{equation}
where H(C) and H(f) are the marginal entropies and H(C, f) is the joint entropy of C and f. H(C), H(f) and H(C, f) are defined by following equations.
\begin{equation}
H(C) = -\sum_{i=1}^K {p(C_i)log_2p(C_i)}
\end{equation}
\begin{equation}
H(f) = -\sum_{x} {p(f=x)log_2p(f=x)}
\end{equation}
\begin{equation}
H(C, ~f) = -\sum_{K} \sum_{n}{p(C, f)log_2p(C,f)}
\end{equation}
where K = number of classes, C = class variable, f = feature set that take discrete values.
\subsubsection{Graph Clustering}
A Graph G = \{E, V, C\} is considered to have a set of edges E = \{$e_{1}, e_{2}, \ldots$\}, vertices V = \{$v_{1}, v_{2}, \ldots$\} and colors C = \{$c_{1}, c_{2}, \ldots$\} where each element of the objects are colored in \{$c_{1}, c_{2}, \ldots$\} and the edges $e_{i}$ are recognised with an disorderly paired vertices \{$v_{1}, v_{2}, \ldots$\} [19] [20] [22]. Vertices which are densely connected or form a sub-graph in a graph can be discovered from graph clustering.

\section{Proposed Methodology}\label{sec2}

The opaque AI systems furnish limited understanding about the input-output relationship. This lack of interpretability is attempted to be addressed with our proposed algorithm. The proposed algorithm Feature Association Map based eXplainability (or FAMeX) comes up with a feature importance based explanation of the system output. The foundation of the FAMeX algorithm is the graph-theoretic concept to model feature association, which is termed as Feature Association Map (FAM). FAM, as shown in Fig. 1, is a graph that has features modelled as vertices and the association between the features represented as edges. There are two facets of this association - one is relevance and the other is redundancy. Relevance signifies the amount of information a particular feature contributes to decide the class value in context of a supervised learning problem. So more information a particular feature contributes, the more significant or relevant it is. Again, multiple features can contribute information in terms of deciding the class value. But certain features might be contributing similar information. Those features can be regarded as potentially redundant. Potentially redundant features can be can be considered to be less important in context of the decision taken by an AI system.


\begin{figure}[h]
\centering
\includegraphics[width=0.6\textwidth]{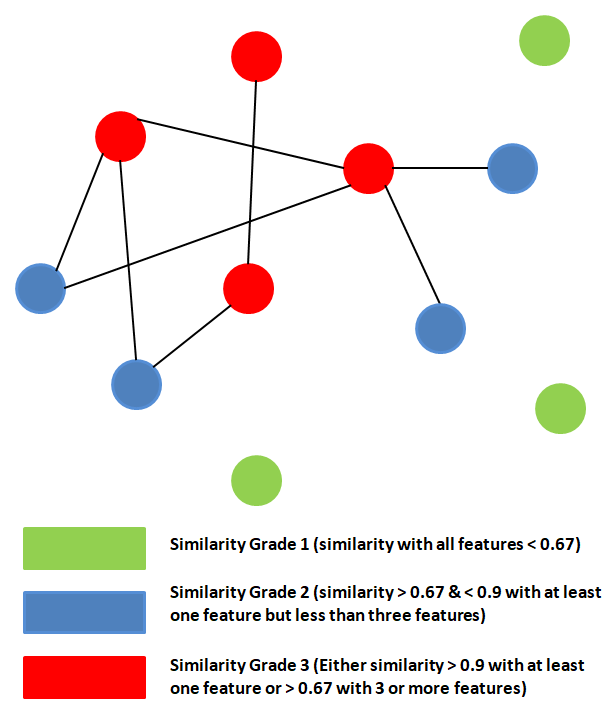}
\captionsetup{justification=centering}
\caption{FAMeX Feature Graph}
\label{fig1}
\end{figure}

In the proposed FAMeX algorithm, the relevance and redundancy level of the features are evaluated. As has been presented in Fig. 1, the overall \textit{feature importance score} is calculated based on two measures - (a) the \textit{similarity score} and (b) the \textit{relevance score}. Based on certain criteria highlighted below, the features obtained from FAM [27, 28] are graded in three levels - Grade 1, Grade 2, and Grade 3. The deciding factors underlying the grades are: 
\begin{itemize}
    \item Grade 1 is assigned to features which have low redundancy i.e. correlation value less than 0.67 with all other features. Feature similarity grade assigned for these features is 1.
    \item Grade 2 is assigned to features which have moderate redundancy i.e. correlation value more than 0.67 but less than 0.9 with at least one feature but less than three features. Feature similarity grade assigned for these features is 2.
    \item Grade 3 is assigned to features which have high redundancy i.e. correlation value more than 0.9 with at least one feature or more than 0.67 with three or more features. Feature similarity grade assigned for these features is 3.
\end{itemize}

Based on this similarity grade, the similarity score is calculated for each feature using the equation 9. 
\begin{equation}
    Similarity\;Score = \frac{{(Feature\;Similarity\;Grade)}^2}{Average\;Feature\;Similarity\;Grade}
\end{equation}
Next, for each feature, feature-to-class mutual information is measured and normalized which is considered as \textit{feature relevance}. Relevance score for each feature is calculated using the equation 10. 
\begin{equation}
    Relevance\;Score = \frac{Feature\;Relevance}{Average\;Feature\;Relevance}
\end{equation}
To calculate feature importance score for each feature, similarity score and the relevance score are integrated in such a way that the score of the feature is higher if the relevance score is higher and the score of the feature is lower if the similarity score is higher. So similarity score, as shown in equation 11, has an inverse relation whereas relevance score has a direct relation with the feature importance score. 
\begin{equation}
    Feature\;Importance\;Score = \frac{Relevance\;Score}{Similarity\;Score}
\end{equation}
Thus, an explanation of the AI system or the learning model is derived based on the feature importance score of the features. The top features are the ones that are the most important features i.e. having highest feature importance score. Those features are expected to be most important in deciding a particular class value. Thus the most important features are the features that decide on the prediction, and these are the features which can be used to explain the model outcome.

\section{Algorithm}\label{sec2}


\begin{algorithm}[H]
\caption{FAMeX Algorithm}
\label{alg:famex}

\begin{algorithmic}[1]

\State \textbf{Input:} N-dimensional data set $D_N$, with original feature set
$F = \{f_1, f_2, \ldots, f_n\}$

\State \textbf{Output:} Feature importance score set
$M = \{m_1, m_2, \ldots, m_n\}$

\State $M_{\text{corr}} \gets |\text{correlation}(D_{N-1})|$

\State $sim \gets M_{\text{corr}}$

\For{each element of $M_{\text{corr}}$}

    \If{$colnum = rownum$}
        \State $M_{\text{corr}}[colnum, rownum] \gets 0$
    \EndIf

    \If{$M_{\text{corr}}[colnum, rownum] \geq 0.67$}
        \State $M_{\text{corr}}[colnum, rownum] \gets 1$
    \Else
        \State $M_{\text{corr}}[colnum, rownum] \gets 0$
    \EndIf

\EndFor

\State $G_{\text{FAM}} \gets \text{adjacency-graph}(M_{\text{corr}})$

\For{each element of $sim$}

    \If{$\text{len}(sim[colnum,rownum] \geq 0.67) = 0$}

        \State $V(G_{\text{FAM}}) \gets \text{color1}$

    \ElsIf{$\text{len}(sim[colnum,rownum] \geq 0.9) \geq 1$ 
    \textbf{or}
    $\text{len}(sim[colnum,rownum] \geq 0.67) \geq 3$}

        \State $V(G_{\text{FAM}}) \gets \text{color3}$

    \Else

        \State $V(G_{\text{FAM}}) \gets \text{color2}$

    \EndIf

\EndFor

\State $\text{vertexcolormap} \gets [color1, color2, color3]$

\State $\text{nodegrade} \gets [1,2,3]$

\For{each feature in $F$}

    \State $simscore \gets
    \frac{(\text{nodegrade})^2}
    {\text{average}(\text{nodegrade})}$

    \State $mi \gets \text{MIClassif}()$

    \State $relscore \gets
    \frac{mi}{\text{mean}(mi)}$

    \State $featimpscore \gets
    \frac{relscore}{simscore}$

\EndFor

\end{algorithmic}
\end{algorithm}
The time complexity of the algorithm is $O(n^2)$ where $n$ is the number of features in a dataset. where n is the number of features in a dataset.
\section{Experiments and Results}\label{sec2}

\subsection{Experimental setup}
Eight benchmark datasets, presented in Table I, have been considered for the experiments. All the datasets are available publicly in the University of California (UCI) Machine Learning repository [29]. These datasets are obtained from various domains to check the variability in the outcomes of the FAMeX and other competing algorithms.

\begin{table}[h]
\caption{Details of the datasets used in experiments}\label{tab1}%
\renewcommand{\arraystretch}{1.3}
\begin{tabular}{@{}lcc@{}}
\toprule
Datasets & No. of Instances & No. of Features\\
\midrule
Wisconsin & 683 & 10 \\ 
ILPD & 580 & 11 \\ 
Pageblocks & 5474 & 11 \\ 
Pima & 769 & 9 \\ 
Apndcts & 107 & 8 \\ 
WineQuality & 1600 & 12 \\ 
WBDC & 570 & 31 \\ 
Vehicle & 847 & 19 \\
\botrule
\end{tabular}
\end{table}

We have used Google Colaboratory platform for programming and Python 3.7 as the programming language.

Results of the proposed FAMeX algorithm have been compared with two benchmark algorithms of Explainable AI, namely PFI and SHAP. This comparative study can provide a better understanding of how the proposed FAMeX algorithm performs in comparison to the existing benchmark algorithms with the same datasets. 10-fold cross-validation has been performed both for the competing algorithms PFI and SHAP as well as for FAMeX. 

\subsection{Evaluation metrics}
To evaluate the performance of the proposed FAMeX algorithm compared to the competing SHAP and PFI algorithms, feature importance based ranking has been considered. To find the relative efficacy of the algorithms, most important features identified by each algorithm have been used to derive a subset of the dataset. Then classification accuracy has been calculated using a standard classifier. It is obvious that the XAI algorithm which is able to select the set of features generating higher classification accuracy based on feature importance, is the best performing algorithm. We have also taken a view by selecting the least important feature. For a given algorithm, it is expected that the data subset generated by most important features should have a significantly higher classification accuracy than the data subset generated by least important features. Few of the robust classifiers, namely Support Vector Machine (SVM) [18, 19], Decision Tree (DT) [20, 21], Random Forest (RF) [22] and Naive Bayes (NB) [23, 24] are used to justify the performance of the algorithms and check how accurate the outcomes are when applied on the datasets. All the experiments have been repeated for 100 iterations.

For the experiments presented in this paper, the top 30\% of the FAMeX features and bottom 30\% of the FAMeX features are extracted and classified. As already stated, it is expected that the classification accuracy for the top 30\% of the features will be significantly better than the bottom 30\% of the features. This will indicate the trustworthiness of the algorithm. This has been applied on each of the competing XAI algorithms too.

\subsection{Results and analysis}
The FAM formation for the WineQuality dataset has been presented in Fig. 2. As can be observed, 5 out of 11 features are represented by green vertices, indicating low similarity / redundancy. Only one feature, fixed.acidity, has similarity grade 3.

\begin{figure}[h]
\centering
\includegraphics[width=0.5\textwidth]{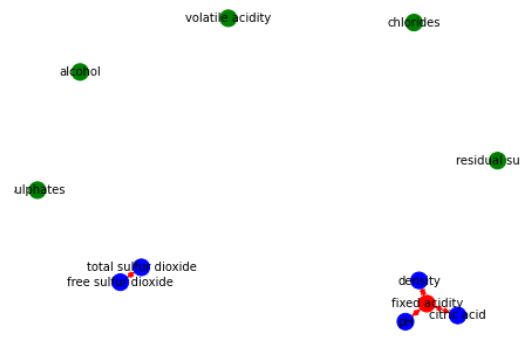}
\captionsetup{justification=centering}
\caption{Association Map for WineQuality dataset}
\label{fig2}
\end{figure}

\begin{table}[h]
\renewcommand{\arraystretch}{1.3}
\caption{Feature importance scores of PFI, FAMeX and SHAP for Wisconsin dataset}\label{tab1}%
\begin{tabular}{@{}lccc@{}}
\toprule
Wisconsin & PFI & FAMEX & SHAP \\ 
\midrule
Bare Nuclei & 0.393 & 1.877 & 5 \\
Clump Thickness & 0.53 & 0.900 & 4 \\
Mitosis & 0.479 & 0.600 & 4 \\
Bland Chromatin & 0.437 & 0.449 & 5 \\ 
Cell Shape & 0.309 & 0.383 & 7 \\ 
Marginal Adhesion & 0.322 & 0.327 & 10 \\ 
Epithelial Size & 0.0916 & 0.304 & 3 \\ 
Cell Size & 0.0324 & 0.268 & 2 \\ 
Normal Nucleoli & 0.217 & 0.239 & 1 \\ 
\botrule
\end{tabular}
\end{table}

\textbf{Table 2} presents a comparison of the feature importance scores of the three algorithms for one of the datasets used in the experiments. From the table it is evident that \textit{Bare Nuclei}, \textit{Clump Thickness} and \textit{Mitosis} have the highest values for feature importance score based on the FAMeX algorithm. For PFI, \textit{Clump Thickness}, \textit{Mitosis} and \textit{Bland Chromatin} have the highest values whereas for SHAP, \textit{Marginal Adhesion} has got the highest value followed by \textit{Cell Shape}, \textit{Bare Nuclei} and \textit{Clump Thickness}. Hence it can be said that along with the existing Explainable AI algorithms PFI and SHAP, FAMeX can extract the set of important features in context of decision making of an AI system. However, since each of the algorithms have come up with a different ranking for the feature importance, albeit some of them appearing in the top of the list for all the algorithms, the efficacy of the XAI algorithm needs to be decided by applying a part of the top / bottom features to determine the accuracy. 

\begin{table}[h]
\renewcommand{\arraystretch}{1.3}
\caption{Accuracy (using SVM) with Top 30\% and Bottom 30\% important features}\label{tab2}
\begin{tabular*}{\textwidth}{@{\extracolsep\fill}lcccccc}
\toprule%
& \multicolumn{3}{@{}c@{}}{Top 30\%} & \multicolumn{3}{@{}c@{}}{Bottom 30\%\footnotemark[2]} \\\cmidrule{2-4}\cmidrule{5-7}%
Dataset & FAMeX & PFI & SHAP & FAMeX & PFI & SHAP \\
\midrule
Wisconsin & 95.41±1\% & 95.22±1\% & 93.93±1\% & 93.72±1\% & 89.58±1\% & 89.24±1\% \\ 
        ILPD & 71.63±1\% & 71.19±1\% & 71.06±1\% & 71.11±1\% & 71.25±1\% & 71.62±1\% \\ 
        Pageblocks & 93.75±2\% & 89.85±2\% & 89.24±2\% & 90.17±2\% & 89.67±2\% & 90.10±2\% \\ 
        Pima & 75.81±1\% & 66.12±1\% & 75.45±1\% & 65.29±1\% & 64.88±1\% & 64.64±1\% \\ 
        Apndcts & 88.67±1\% & 80.5±1\% & 83.09±1\% & 83.54±1\% & 85.68±1\% & 81.22±1\% \\ 
        WineQuality & 55.62±1\% & 49.98±1\% & 49.01±1\% & 44.10±1\% & 55.41±1\% & 46.11±1\% \\ 
        WBDC & 91.75±2\% & 85.22±2\% & 85.35±2\% & 62.98±2\% & 92.85±2\% & 92.51±2\% \\ 
        Vehicle & 43.56±2\% & 40.69±2\% & 41.07±2\% & 42.7±2\% & 37.20±2\% & 36.77±2\% \\ 
        \textbf{Average} & \textbf{77.03±1\%} & \textbf{72.35±1\%} & \textbf{73.53±1\%} & \textbf{69.20±1\%} & \textbf{72.07±1\%} & \textbf{71.53±1\%} \\
\botrule
\end{tabular*}
\end{table}
Let's first start with the results of the SVM classifier presented in \textbf{Table 3}. As can be observed, for all the datasets FAMeX generates better classification accuracy than the competing PFI and SHAP algorithms. On an overall perspective with the top 30\% features, FAMeX generates an average accuracy of 77\% compared 72.35\% by PFI and 73.53\% by SHAP. On the other hand, for most of the datasets, the classification accuracy generated by top 30\% features identified by FAMeX is significantly higher than the accuracy generated by bottom 30\% features. For a few datasets like Pima, Wine and WBDC, the difference is more than 10\%.

\begin{table}[h]
\renewcommand{\arraystretch}{1.3}
\caption{Accuracy (using RF) with Top 30\% and Bottom 30\% important features}\label{tab2}
\begin{tabular*}{\textwidth}{@{\extracolsep\fill}lcccccc}
\toprule%
& \multicolumn{3}{@{}c@{}}{Top 30\%} & \multicolumn{3}{@{}c@{}}{Bottom 30\%\footnotemark[2]} \\\cmidrule{2-4}\cmidrule{5-7}%
Dataset & FAMeX & PFI & SHAP & FAMeX & PFI & SHAP \\
\midrule
Wisconsin & 94.38±1\% & 93.77±1\% & 94.29±1\% & 93.52±1\% & 89.66±1\% & 89.04±1\% \\
        ILPD & 68.19±1\% & 69.98±1\% & 68.13±1\% & 65.76±1\% & 63.87±1\% & 68.53±1\% \\ 
        Pageblocks & 97.25±2\% & 90.71±2\% & 90.08±2\% & 52.26±2\% & 90.34±2\% & 91.22±2\% \\ 
        Pima & 73.49±1\% & 59.44±1\% & 67.37±1\% & 63.64±1\% & 58.50±1\% & 61.42±1\% \\ 
        Apndcts & 85.82±1\% & 78.45±1\% & 78.54±1\% & 84.40±1\% & 83.31±1\% & 79.0±1\% \\ 
        WineQuality & 65.16±1\% & 52.94±1\% & 52.17±1\% & 56.75±1\% & 57.35±1\% & 49.86±1\% \\ 
        WBDC & 92.26±2\% & 82.39±2\% & 83.42±2\% & 70.43±2\% & 90.35±2\% & 90.21±2\% \\ 
        Vehicle & 64.74±2\% & 44.1±2\% & 44.52±2\% & 60.29±2\% & 38.25±2\% & 53.50±2\% \\ 
        \textbf{Average} & \textbf{80.16±1\%} & \textbf{71.47±1\%} & \textbf{71.82±1\%} & \textbf{68.38±1\%} & \textbf{71.45±1\%} & \textbf{72.85±1\%} \\
\botrule
\end{tabular*}
\end{table}

For Random Forest classifier, as shown in \textbf{Table 4}, for most of the datasets FAMeX generates better classification accuracy than the competing PFI and SHAP algorithms. On an overall perspective with the top 30\% features, FAMeX generates an average accuracy of 80.16\% compared 71.47\% by PFI and 71.82\% by SHAP. For all the datasets, the classification accuracy generated by top 30\% features identified by FAMeX is significantly higher than the accuracy generated by bottom 30\% features. For a few datasets like Pageblocks, Pima, Wine and WBDC datasets, the difference is more than 10\%.

\begin{table}[h]
\renewcommand{\arraystretch}{1.3}
\caption{Accuracy (using NB) with Top 30\% and Bottom 30\% important features}\label{tab2}
\begin{tabular*}{\textwidth}{@{\extracolsep\fill}lcccccc}
\toprule%
& \multicolumn{3}{@{}c@{}}{Top 30\%} & \multicolumn{3}{@{}c@{}}{Bottom 30\%\footnotemark[2]} \\\cmidrule{2-4}\cmidrule{5-7}%
Dataset & FAMeX & PFI & SHAP & FAMeX & PFI & SHAP \\
\midrule
Wisconsin & 95.14±1\% & 95.04±1\% & 93.32±1\% & 94.25±1\% & 85.62±1\% & 88.98±1\% \\ 
        ILPD & 70.72±1\% & 50.48±1\% & 69.07±1\% & 53.69±1\% & 68.36±1\% & 49.98±1\% \\ 
        Pageblocks & 94.08±2\% & 78.21±2\% & 87.50±2\% & 92.97±2\% & 78.35±2\% & 90.45±2\% \\ 
        Pima & 75.55±1\% & 67.98±1\% & 74.92±1\% & 65.62±1\% & 65.55±1\% & 66.46±1\% \\ 
        Apndcts & 87.68±1\% & 66.59±1\% & 83.77±1\% & 84.31±1\% & 84.99±1\% & 65.36±1\% \\ 
        WineQuality & 56.81±1\% & 49.31±1\% & 48.3±1\% & 41.93±1\% & 55.83±1\% & 40.39±1\% \\ 
        WBDC & 91.50±2\% & 82.13±2\% & 82.39±2\% & 76.42±2\% & 90.11±2\% & 90.78±2\% \\ 
        Vehicle & 43.53±2\% & 31.09±2\% & 31.13±2\% & 45.11±2\% & 53.5±2\% & 38.27±2\% \\ 
        \textbf{Average} & \textbf{76.88±1\%} & \textbf{65.10±1\%} & \textbf{71.3±1\%} & \textbf{69.28±1\%} & \textbf{72.78±1\%} & \textbf{66.33±1\%} \\
\botrule
\end{tabular*}
\end{table}

As presented in \textbf{Table 5}, for Naive Bayes classifier, for all the datasets FAMeX generates better classification accuracy than the competing PFI and SHAP algorithms. On an overall perspective with the top 30\% features, FAMeX generates an average accuracy of 76.88\% compared 65.1\% by PFI and 71.3\% by SHAP. For all the datasets excluding Vehicle, the classification accuracy generated by top 30\% features identified by FAMeX is significant7ly higher than the accuracy generated by bottom 30\% features. For a few datasets like ILPD, Wine and WBDC datasets, the difference is more than 10\%.

\begin{table}[h]
\renewcommand{\arraystretch}{1.3}
\caption{Accuracy (using DT) with Top 30\% and Bottom 30\% important features}\label{tab2}
\begin{tabular*}{\textwidth}{@{\extracolsep\fill}lcccccc}
\toprule%
& \multicolumn{3}{@{}c@{}}{Top 30\%} & \multicolumn{3}{@{}c@{}}{Bottom 30\%\footnotemark[2]} \\\cmidrule{2-4}\cmidrule{5-7}%
Dataset & FAMeX & PFI & SHAP & FAMeX & PFI & SHAP \\
\midrule
Wisconsin & 94.38±1\% & 93.48±1\% & 93.86±1 & 93.67±1\% & 89.16±1\% & 88.82±1\% \\ 
        ILPD & 70.28±1\% & 66.36±1\% & 65.68±1\% & 64.93±1\% & 62.56±1\% & 66.90±1\% \\ 
        Pageblocks & 96.46±2\% & 90.72±2\% & 90.12±2\% & 91.65±2\% & 90.70±2\% & 91.37±2\% \\ 
        Pima & 68.91±1\% & 57.48±1\% & 65.72±1\% & 59.79±1\% & 59.06±1\% & 61.50±1\% \\ 
        Apndcts & 81.41±1\% & 75.86±1\% & 70.95±1\% & 79.0±1\% & 77.09±1\% & 75.59±1\% \\ 
        WineQuality & 61.04±1\% & 53.17±1\% & 52.72±1\% & 54.42±1\% & 56.75±1\% & 49.80±1\% \\ 
        WBDC & 90.39±2\% & 79.87±2\% & 80.97±2\% & 72.14±2\% & 88.38±2\% & 89.07±2\% \\ 
        Vehicle & 61.38±2\% & 43.41±2\% & 43.29±2\% & 57.33±2\% & 53.91±2\% & 53.67±2\% \\ 
        \textbf{Average} & \textbf{78.03±1\%} & \textbf{75.11±1\%} & \textbf{70.41±1\%} & \textbf{71.61±1\%} & \textbf{72.20±1\%} & \textbf{70.84±1\%} \\ 
\botrule
\end{tabular*}
\end{table}

As shown in \textbf{Table 6}, the results are consistent for Decision Tree classifier too. For all datasets FAMeX generates better classification accuracy than the competing PFI and SHAP algorithms. On an overall perspective with the top 30\% features, FAMeX generates an average accuracy of 78\% compared to 75.11\% by PFI and 70.41\% by SHAP. For all the datasets, the classification accuracy generated by top 30\% features identified by FAMeX is significantly higher than the accuracy generated by bottom 30\% features.

\begin{table}[h]
\renewcommand{\arraystretch}{1.3}
\caption{Average Accuracy (all classifiers) with Top 30\% and Bottom 30\% important features}\label{tab2}
\begin{tabular*}{\textwidth}{@{\extracolsep\fill}lcccccc}
\toprule%
& \multicolumn{3}{@{}c@{}}{Top 30\%} & \multicolumn{3}{@{}c@{}}{Bottom 30\%\footnotemark[2]} \\\cmidrule{2-4}\cmidrule{5-7}%
Dataset & FAMeX & PFI & SHAP & FAMeX & PFI & SHAP \\
\midrule
Support Vector Machine & 77.03±1\% & 72.35±1\% & 73.53±1\% & 69.20±1\% & 72.07±1\% & 71.53±1\% \\ 
        Random Forest & 80.16±1\% & 71.47±1\% & 71.82±1\% & 68.38±1\% & 71.45±1\% & 72.85±1\% \\ 
        Naive Bayes & 76.88±1\% & 65.10±1\% & 71.3±1\% & 69.28±1\% & 72.78±1\% & 66.33±1\% \\ 
        Decision Tree & 78.03±1\% & 75.11±1\% & 70.41±1\% & 71.61±1\% & 72.20±1\% & 70.84±1\% \\ 
\botrule
\end{tabular*}
\end{table}

The summary of results of all experiments have been presented in \textbf{Table 7}. As can be observed, FAMeX demonstrates consistently better results than the competing algorithms with all 4 classifiers. Also, the results for FAMeX with top 30\% features is significantly better than the ones with bottom 30\% features. This establishes the efficacy of FAMeX in coming up with the feature importance scores consistent with the prediction accuracy. However, as can be observed from the results, PFI and SHAP is not able to demonstrate such consistent results. In fact few results show that the top 30\% features gives a lower accuracy than bottom 30\% features.  
\begin{samepage}
\subsection{GUI based tool}
A Graphical User Interface (GUI) based tool has been developed which integrates the implemention of the FAMeX algorithm. The tool allows the user to browse a dataset. The tool generates the similarity score, relevance score and feature importance score of the features of the dataset selected along with the classification accuracy for the 4 standard classifiers - SVM, RF, NB and DT. The percentage of top or bottom features can be taken as user input. This tool is platform-independent. This tool will assist in understanding the important features of a particular dataset in context of a prediction. This prototypic tool has been uploaded in Github repository \url{https://github.com/Sayantanighosh17github/FAMeX-Tool}, so that it can be used by the AI community.

\begin{figure}[h]
\centering
\includegraphics[width=0.6\textwidth]{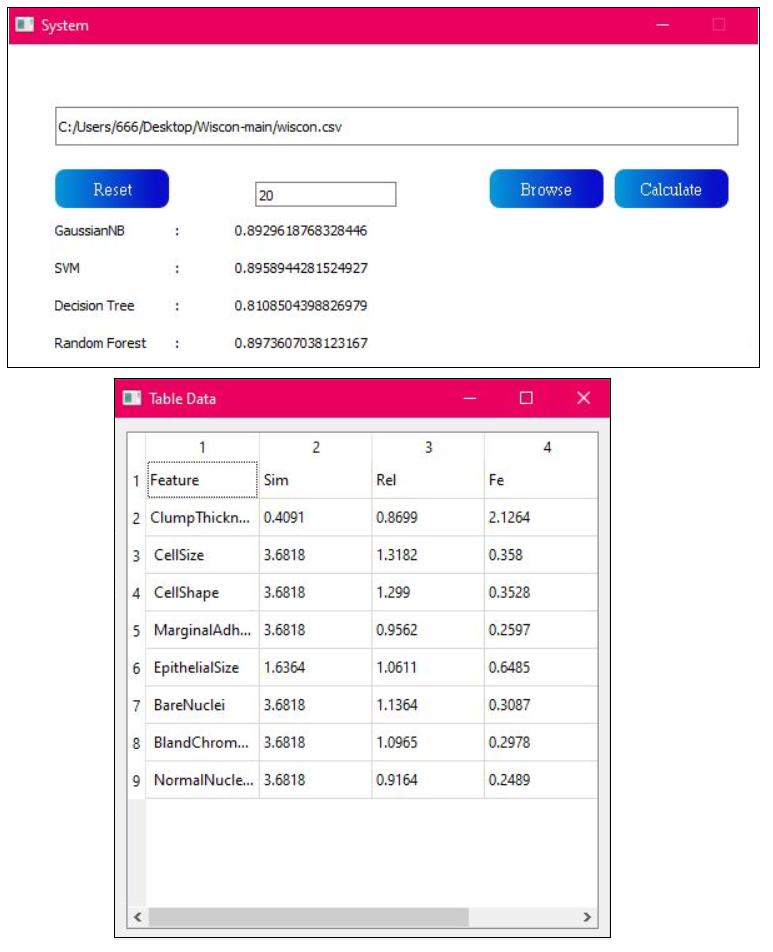}
\captionsetup{justification=centering}
\caption{FAMeX Feature Detector Tool}
\label{fig2}
\end{figure}
\end{samepage}

\section{Conclusion}
In this paper, a new algorithm FAMeX has been proposed for explaining AI systems in terms of most important features. The algorithm has been validated to generate greater accuracy for the top 30\% features and comparatively lesser for the bottom 30\% features. It is also found to outperform two competing XAI algorithms - PFI and SHAP. The experimental results are promising enough to establish the efficacy of FAMeX ahead of PFI and SHAP algorithms. As a future work, the FAMeX algorithm can be applied on more complex and large datasets and compared with more state-of-the-art XAI algorithms.  

\section{Declarations}

\subsection{Availability of data and material}
The data used in this study are from open-source repositories and publicly available sources, as detailed in \textbf{Section 6.1} of this manuscript. All information and datasets referenced are accessible to the public and do not require additional permissions for access.

\subsection{Funding and/or Conflicts of Interests/Competing Interests} 
The authors declare that they have no competing interests or conflicts of interest in connection with this work. No specific funding was received for this study or any related activities.


\end{document}